\documentclass[letterpaper, 10 pt, conference]{ieeeconf}  

\IEEEoverridecommandlockouts                              

\usepackage{subfigure}

\usepackage{eso-pic}					
\usepackage{graphicx}
\usepackage{tikz}
\usetikzlibrary{fadings}
\usepackage{setspace}
\usepackage{amsmath}
\usepackage{mathdots}
\usepackage{yhmath}
\usepackage{siunitx}
\usepackage{array}
\usepackage{gensymb}
\usepackage{amssymb}
\usepackage{mathtools}              
\usepackage[printonlyused]{acronym}
\usepackage{cancel}
\usepackage{color}
\usepackage{multirow}
\usepackage{booktabs}
\usepackage{textcomp}               
\usepackage{svg}        
\usepackage{hyperref}
 
\DeclareMathOperator{\EX}{\mathbb{E}}
\DeclareMathOperator*{\argmax}{argmax}

\usepackage{algorithm}
\usepackage{algpseudocode}
\usepackage{cleveref}
\usepackage{subcaption}

\title{\LARGE \bf
Efficient Non-Myopic Layered Bayesian Optimization For Large-Scale Bathymetric Informative Path Planning
}

\author{Alexander Kiessling$^{1}$, Ignacio Torroba$^{1,2}$, Chelsea Rose Sidrane$^{1,3}$, Ivan Stenius$^{2,3}$ \\ Jana Tumova$^{1,3}$, John Folkesson$^{1}$
\thanks{The authors are with $^{1}$ the \href{https://www.kth.se/is/rpl/division-of-robotics-perception-and-learning-1.779439}{Division of Robotics, Perception and Learning}, $^2$ the \href{https://www.kth.se/en/tekmek/forskargrupper/marina-system/marina-system-1.1238191}{Division of Naval Architecture}, $^3$  \href{https://www.digitalfutures.kth.se/}{Digital Futures} at KTH Royal Institute of Technology, SE-100 44 Stockholm, Sweden.
{\tt\small \{akie, torroba, chelse, stenius, tumova, johnf\}@kth.se}}
}

\begin{document}

\maketitle
\thispagestyle{empty}
\pagestyle{empty}

\begin{abstract}
Informative path planning (IPP) applied to bathymetric mapping allows AUVs to focus on feature-rich areas to quickly reduce uncertainty and increase mapping efficiency. Existing methods based on Bayesian optimization (BO) over Gaussian Process (GP) maps work well on small scenarios but they are short-sighted and computationally heavy when mapping larger areas, hindering deployment in real applications. To overcome this, we present a 2-layered BO IPP method that performs non-myopic, real-time planning in a tree Search fashion over large Stochastic Variational GP maps, while respecting the AUV motion constraints and accounting for localization uncertainty. Our framework outperforms the standard industrial lawn-mowing pattern and a myopic baseline in a set of hardware in the loop (HIL) experiments in an embedded platform over real bathymetry.
\end{abstract}

\section{INTRODUCTION}
\label{sec:introduction}
While the oceans cover more than 70\% of the earth's surface, currently global coverage of bathymetric sonar data encompasses less than 9\% of the seafloor \cite{mayer2018nippon}. It is estimated that more than 900 ship years of surveying are needed to obtain complete sonar coverage of the world seabed \cite{weatherall2015new}. 

The de facto industrial approach to bathymetric surveying with multibeam echosounders (MBES), both with ship hull-mounted equipment or autonomous underwater vehicles (AUVs), consists of lawn-mowing (LM) patterns at a constant depth \cite{galceran2013survey}. 
Such patterns are designed offline during the survey planning stage and are standard because they aim at guaranteeing uniform area coverage and offer a deterministic mission duration. 
The shortcomings of such patterns, however, are three-fold: i) equal distribution of the survey time over unequally distributed topography ultimately means that the same resources are devoted to mapping both feature-rich and uninteresting areas. ii) abrupt changes in the bathymetry might result in lower-resolution areas or gaps in the survey, which are only detected \textit{a posteriori} and thus require a second mission, and iii) the size of the survey area that can be mapped is directly proportional to the mission time and vehicle speed, regardless of the topography of the area. Ideally, however, we would like the mission time to be proportional to the topography of the area, such that uninteresting areas can be mapped faster, thus reducing survey time and increasing efficiency.
\begin{figure}[!t]
    \centering
    \includegraphics[width=1\linewidth]{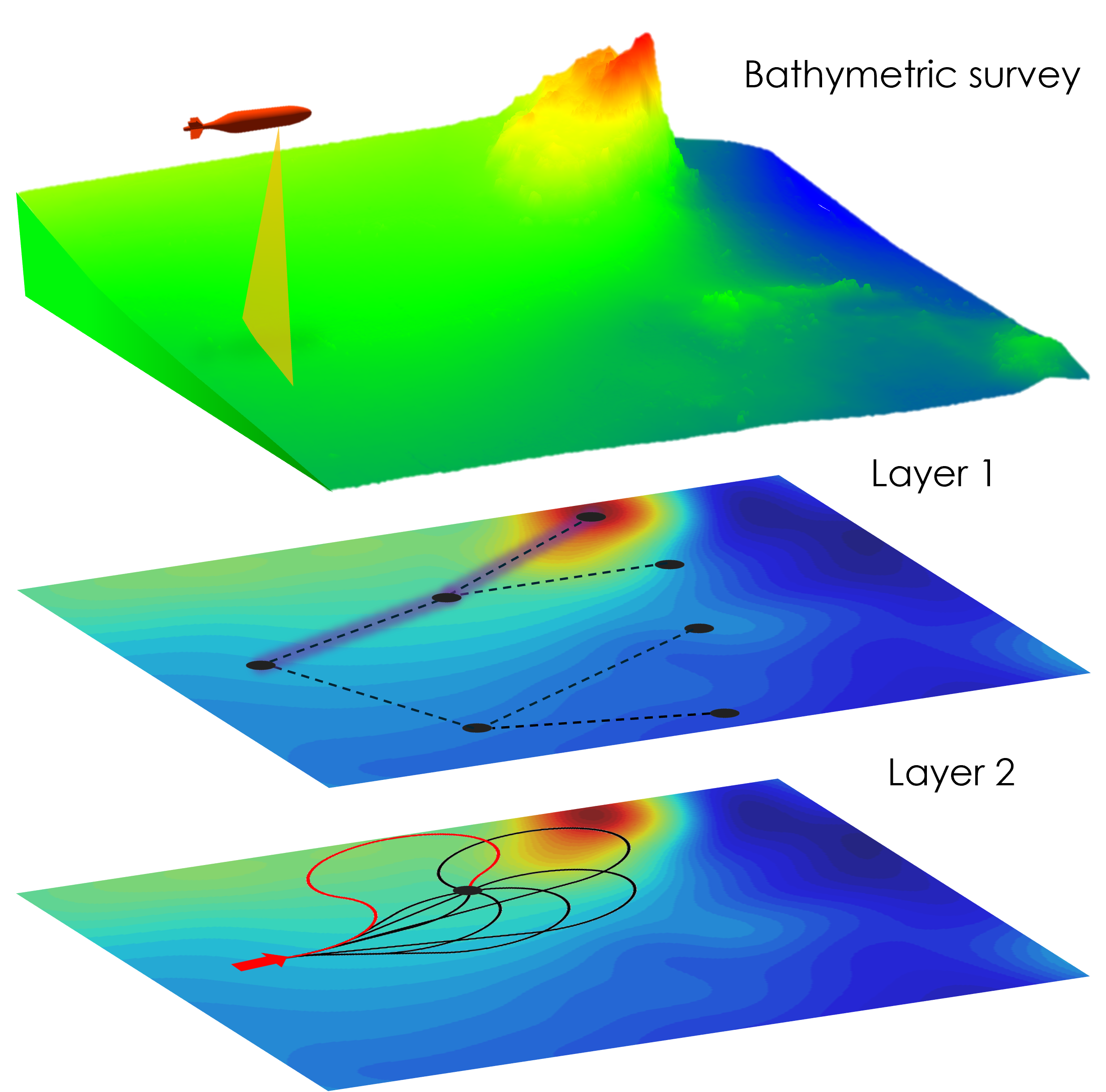}
    \caption{Our 2-layer IPP: the latest AUV pose and the MBES data collected so far are used to regress online a SVGP-UI model of the bathymetry (layer 1) over which a non-myopic BO stage selects the optimal next viewpoint. A feasible path is then selected on a second BO stage (layer 2) by maximizing the expected information collected.}
    \label{fig:system_overview}
\end{figure}
Informative path planning for autonomous exploration has proved to be a tool capable of ameliorating the problems above \cite{zhu2021online, viseras2019robotic}; exchanging the current approach of full coverage survey paths to instead generating paths online, based on collected data. This allows for targeting more informative areas for further collection, and as such areas of high interest can be mapped more quickly, and at a higher resolution. However, underwater bathymetric mapping entails particular challenges that need to be addressed for the successful deployment of an IPP framework in an AUV survey, namely: i) the size of the mapping area is often very large, ii) localization uncertainty increases rapidly, and iii) there are very limited computational resources onboard AUVs. 


To overcome these challenges, we present a two-layered Bayesian optimization (BO) IPP method that performs non-myopic, real-time planning over large areas while respecting the AUV motion constraints and accounting for localization uncertainty. Our contributions with respect to the problems mentioned before are as follow:

\begin{itemize}
    \item Substantially reduced survey time compared to LM patterns for equivalent map quality.
    \item Survey scalability with limited resources via online SVGP regression with uncertain inputs (UIs) for mapping under localization uncertainty.
    \item Real-time performance via batch acquisition functions for BO, enabling more efficient non-myopic planning.
\end{itemize}




\section{RELATED WORK}
Informative Path Planning has been used for offline planning in robotic information gathering tasks~\cite{hollinger2014sampling}, but online re-planning is more robust to localization uncertainty and can incorporate updated map information. 
Some approaches perform online IPP with a pure 
uncertainty reduction
objective~\cite{zhu2021online, viseras2019robotic}, but this does not allow for exploration of interesting features that are only revealed during traversal.
Instead, Bayesian Optimization (BO) \cite{frazier2018tutorial} allows the vehicle's path to be optimized according to an acquisition function consisting of an information gathering objective derived from map uncertainty as well as an objective encouraging exploration of interesting features. 
In addition, BO simplifies the evaluation of potentially expensive objective functions 
through the use of a surrogate model. 

Several BO-based IPP works exist in the literature. 
In \cite{bai2016information}, a BO-based IPP framework is presented that applies GP regression as a surrogate model to optimize a complex information gain objective in a discrete, 2D setting. 
Similarly, the work of~\cite{marchant2014bayesian} employs a layered approach with two separate GPs; one to model the environment and a second to model the quality of potential paths for a vehicle in an exploration task. 
We apply an equivalent decoupled approach in our work, first finding an optimal next viewpoint for the vehicle and then optimizing the path to that point.
However, in our path search we incorporate a more realistic acquisition function to reflect that MBES sensors measure swaths, which are represented as area integrals. Furthermore, we instead apply Stochastic Variational Gaussian Process (SVGP) \cite{hensman2013gaussian} to decouple the regression complexity from the survey size.

BO can provide reasonable performance for small-scale problems but it does not take into account the value of future information.
In \cite{morere2017sequential} and \cite{morere2018continuous}, a non-myopic BO approach is used to compensate for this. In these works, the IPP problem is posed as a continuous Partially Observable Markov Decision Process (POMDP) \cite{sondik1971optimal} to be solved with Monte Carlo Tree Search (MCTS) \cite{coulom2006efficient}. To expand the search tree with actions in the continuous space,  \cite{morere2017sequential} applies BO to sequentially search for optimal candidates. This approach is termed Continuous Belief Tree Search (CBTS), and has seen use in planning for the underwater domain in \cite{duecker2021embedded}. 
Our proposed method is non-myopic as well and we also solve the resultant continuous POMDP with tree search. 
However, to expand the search tree we instead use the reparametrization trick of acquisition functions as described in \cite{wilson2017reparameterization} to achieve optimization with parallel candidates, drastically reducing the number of world representations (SVGPs) needed.

Most path planning methods assume perfect pose information, but this assumption is particularly brittle in underwater domains due to the absence of GPS which creates significant localization uncertainty. 
To account for localization drift during surveys, the vehicle uncertainty can be propagated to the Gaussian process model of the environment. \cite{ghaffari2019sampling} and \cite{popovic2020informative} achieve this through the use of uncertain inputs (UIs) in their IPP implementations, showcasing better performance when compared with deterministic inputs. However, their uncertainty propagation approach does not scale well with the number of training points and their dimensions, limiting the size of the areas that can be surveyed.
To circumvent these shortcomings we instead combine the UI construction from \cite{torroba2022fully}, where a Monte-Carlo approximation of the UI distributions is used, together with the online SVGP mapping framework from \cite{torroba2023online}. Thus, our IPP framework can manage surveys of millions of samples online while accounting for vehicle localization uncertainty.


\section{METHODOLOGY}
The informative path planning problem is defined as selecting a path $\psi$ which approximately maximizes an information metric $I(\psi)$ with respect to a cost constraint $c(\psi) < C$ as follows:
\begin{equation}
    \label{eq:ipp}
    \psi^* = \argmax_{\psi \in \Psi} \frac{I(\psi)}{c(\psi)}, \;\;\;\;
    \mathrm{s.t \; } \sum_\psi c(\psi) \leq C.
\end{equation}

We plan path segments $\psi_i$ online in a receding horizon manner with the maximum travelled distance as our budget $C$.
We constrain our planning problem to a fixed surveying depth, as is standard underwater \cite{galceran2013survey}, and thus each path segment $\psi_i$ has starting and ending viewpoints given by $s_i=[x_i, y_i, \theta_i]$ and $s_{i+1}=[x_{i+1}, y_{i+1}, \theta_{i+1}]$.

A schematic representation of our framework to solve Eq. \ref{eq:ipp} can be seen in Fig.~\ref{fig:system_overview}. In a first stage, a non-myopic BO step over an SVGP-UI model of the seabed is carried out to find the position components of the next best viewpoint, $s_{i+1}^{x, y}$. 
Once $s_{i+1}^{x, y}$ has been selected, the optimal vehicle path $\psi_i^*$ to reach $s_{i+1}^{x, y}$ from current point $s_{i}$ is sought. By parameterizing the AUV paths as Dubins paths  \cite{dubins1957curves}, $\psi_i^*$ is fully determined by choosing $\theta_{i+1}$ given $s_i$ and $s_{i+1}^{x, y}$ and we denote a path segment parameterized by ending angle $\psi_i(\theta_{i+1})$ . 
Therefore, the second BO step seeks to find the optimal path segment $\psi_i^*(\theta_{i+1})$ by maximizing the MBES sensor data collected along the path. 
Finally, as the vehicle traverses the selected path and collects more MBES data, the planning for the next location resumes in an receding horizon fashion, when the remaining path is less than a predefined distance.  
In this section, the individual components of the framework are presented in more detail.


\subsection{Scalable Bathymetry Regression Under Uncertainty}
Bathymetric surveying consists of building a map of a seabed area $E$ from MBES data collected with an AUV. Assuming zero-mean Gaussian noise with uncertainty $W$, the motion of the AUV can be modelled as $s_{i} \sim \mathcal{N}(m(s_{i-1}, u_{i}), W)$, where $m()$ are the AUV motion equations and $u_i$ the control inputs.
The MBES pings can be parameterized as $B_{i} \sim \mathcal{N}(h(s_i, E_i), Q)$, where $B_i = \{b_{i}\}_{i=1}^{n}$ is a ping containing $n$ 3D beams $b_i = [x, y, z]$, $h()$ is the MBES measurement model and $Q$ parameterizes the measurements uncertainty. $E_i = \{e_{j}\}_{j=1}^{n}$ describes the patches of seabed, in the map frame, where the beams hit the ground at $i$. 
Modelling the seabed as an unknown continuous function of the form $f: X \to Z$ from which the MBES beams are noisy samples, GP regression can be used to approximate the original $f(X)$ from the 3D MBES data, with $X=[b_j^x,b_j^y]$ and $Z=b_j^z$ being the depth of the location. 
This would take the form $f(X) \sim \mathcal{GP}\big(\mu(X), k(X, X') \big)$ where $\mu(X)$ and $k(X, X')$ are the mean and covariance functions.

As introduced in Sec. \ref{sec:introduction}, two major impediments towards achieving embedded BO-based IPP for bathymetric surveying with AUVs are the large scale of the seabed areas to map, comprising millions of samples, and the localization uncertainty of AUVs.
In order to overcome this and regress surveys of millions of beams online while accounting for AUV localization uncertainty on the first layer of BO, we combine the approaches from \cite{torroba2022fully} and \cite{torroba2023online} to learn SVGP with uncertaint inputs (UIs) maps online. 
Regressing $f$ as such amounts to minimize the following SVGP ELBO with minibatch stochastic gradient descent (SGD), as in \cite{torroba2023online}:
\begin{equation}
\label{eq:elbo_mini_svgp}
    \mathcal{\hat{L}} = \frac{N_t}{M} \sum^M_{j}\mathbb{E}_{q(u) p(f | u)} \left[\ln p(z_j \mid \hat{f_j}))\right] - \mathrm{KL}\left[q(u) \mid\mid p(u)\right]
\end{equation}
where $u$ are the inducing points,  $q(u)$ is a multivariate normal variational term, $M$ is the minibatch size and $N_t$ are the beams collected so far on the survey.
The difference between Eq. \ref{eq:elbo_mini_svgp} and that in \cite{torroba2023online} is that $\hat{f_j} = f(\hat{x}_j)$, with $\hat{x}_j  \sim \mathcal{N}(x_j, \Omega_j)$ being drawn in a Monte Carlo fashion from the beams UIs, where $\Omega_j$ is constructed following the method in \cite{torroba2022fully}. Due to space constraints, the reader is referred to \cite{torroba2022fully, torroba2023online} for a more in depth explanation of Eq. \ref{eq:elbo_mini_svgp}.


\subsection{Layer 1: BO-based Next Best Viewpoint}
Our first layer of BO directly utilizes the SVGP-UI as a surrogate $g()$ of the seabed regressed from the MBES collected so far to select a $s_{i+1}^{x, y}$ that maximizes information gathered. We apply the commonly used upper confidence bound (UCB) acquisition function \cite{auer2002using} for this, 
\begin{equation} 
\mathrm{UCB}(g(s^{x, y})) = \mu(s^{x, y}) + \sqrt{\beta}\, \sigma(s^{x, y}),
\label{eq:ucb}
\end{equation}
where $\mu(s^{x, y})$ and $\sigma(s^{x, y})$ are the SVGP posterior mean and variance at the AUV location $s^{x, y}$, respectively, and $\beta$ is used to balance exploration and exploitation.
This results in a BO layer whose objective function is as follows:
\begin{equation}
    s^{x, y*} = \argmax\limits_{s^{x, y}} \mathrm{UCB}[g(s^{x, y})]
    \label{eq:bo}
\end{equation}
However, the myopic BO objective function in Eq.~\eqref{eq:bo} can yield sub-optimal global results on surveys over large areas due to its short-sightedness \cite{marchant2014sequential}. To overcome this, we pose our optimization problem as a continuous Partially Observable Markov Decision Process (POMDP) and apply efficient tree search to solve it in a non-myopic manner.

\subsubsection{Non-Myopic Tree Search}
We define our POMDP as a tuple $(\mathcal S, \mathcal A, \mathcal T, \mathcal O, r, \gamma=1)$ with a continuous state space $s = [x,y,\theta]^T$, a parameterized continuous action space defined by Dubins paths $a = \psi(s_1, s_2)$, a transition function defined by the AUV motion model $\mathcal T = m()$, an observation function given by the MBES measurement model $\mathcal O = h()$, and a reward function $r = V()$, which will be defined in detail below.

To solve the POMDP problem, we perform a tree search similar to MCTS. Standard MCTS consists of four components, namely selection, backpropagation, rollout and expansion~\cite{sutton2018reinforcement}.
For selection and backpropagation we adopt flat UCT \cite{coquelin2007bandit}, which backpropagates maximum values through the search tree, rather than performing a Bellman backup.
This approach retains the adaptability of standard UCT from \cite{kocsis2006bandit}, improves the regret bound in worst cases where UCT is overly optimistic, 
and is suitable as our method specializes in searching for features of interest.  For the rollout step, we use the expected value of random samples of uncertainty $\EX(\sigma(X))$, with the samples $X$ distributed uniformly around a viewpoint. This ensures efficient rollouts that prioritize unexplored viewpoints.

As action selection in a continuous space is not tractable in standard MCTS \cite{papadimitriou1987complexity}, the expansion step requires special attention. 
The expansion step acts as discretization of the action space, and we use BO to select a set of discrete optimal actions as in \cite{morere2018continuous}. 
However, we develop a distinctly different concept in our method to increase computational performance, introduced below.

\subsubsection{Batch BO for Efficient Tree Search}
The approach in \cite{morere2018continuous} discretizes the action space by sequentially solving Eq.\ref{eq:bo} and updating the corresponding surrogate $g()$ through Eq. \ref{eq:elbo_mini_svgp} with each iteration. This involves training a large number of SVGP representations in order to expand the tree, with consequences on performance. At any depth of a search tree, to yield $q$ candidates the discretization step requires $q$ GPs to be trained, due to conditioning. 
To overcome this, we expand the search tree employing the q-batch UCB function instead, in Eq. \ref{eq:qucb}, derived from applying the reparameterization trick to the original UCB, as described in \cite{wilson2017reparameterization}:
\begin{equation}
\mathrm{qUCB} \approx \frac{1}{N} \sum_{n=1}^N\max\limits_{j=1,...,q}(\mu_j + |L a_n|), \;\; a_n \sim \mathcal{N} (0, I),
\label{eq:qucb}
\end{equation}
where $\mu_j = \mu(s_{j}^{x, y})$, $\Sigma \in \mathbb{R}^{q \times q}$ is the diagonal covariance matrix constructed from the $q$ values $\sigma(s_q^{x, y})$ and $LL^T = \Sigma \frac{\beta \pi}{2}$ is a Cholesky decomposition.
Whilst using Eq. \ref{eq:ucb} to expand the search tree into $q$ child nodes requires sequentially conditioning on each of the nodes in turn, Eq. \ref{eq:qucb} yields $q$ near optimal solution candidates, which allows us to expand the search tree without having to condition on any of these nodes. Thus, substituting Eq. \ref{eq:qucb} in the original Eq. \ref{eq:bo} we effectively reduce the number of SVGP-UI nodes needed to expand a branch of the search tree to a single GP, regardless of the size of $q$. This pairs particularly well with our efficient rollout method, as the variance samples $\sigma(X)$ do not rely on conditioning. These rollout samples can instead be sampled from the parent GP representation in the search tree, amortizing conditioning of the frontier. 



\subsection{Layer 2: Constrained Path Optimization}
The second BO stage receives as inputs $s_{i+1}^{x, y*}$ from the tree search above and the current AUV location $s_i$ from the vehicle's localization module $m()$, and outputs the optimal path $\psi_i$ between them in terms of information gathered with the MBES.
We have chosen to parameterize this 2D path as a Dubins path for two reasons: i) given the minimum turning radius of the AUV, Dubins will produce a feasible optimal path between $s_{i+1}^{x, y*}$ and $s_{i}$, ii) said path will be uniquely defined by the heading of the AUV upon arrival, $\psi(\theta^*_{i+1})$, being therefore very fast to optimize. This fast 1D optimization seeks to maximize the value of the information gathered along a path by the MBES, namely $V\big[\psi(\theta)\big]$. The value of a path $V$ can be computed as follows

\begin{equation}
    V(\psi) = \iint_\mathbb{E(\psi)} UCB(g(X)) \; dE,
    \label{eq:value}
\end{equation}
where $E(\psi)$ is the 2D region of the seabed captured by the MBES swath along the path $\psi$. In order to solve this integral efficiently we apply MC approximation and sample from the set of MBES beams $X$ collected from $E(\psi)$ to approximate $V$.
Given $l: \theta \to V$, we can regress $l$ with a GP and find its optimal value through a second layer of BO. This can be formally expressed as $\theta^* = \argmax_\theta V\big[\psi(\theta)\big]$, which together with Eq. \ref{eq:bo} fully solves Eq. \ref{eq:ipp}.
With the search space limited to $\theta \in [-\pi, \pi]$, we apply exact GP regression on this second BO stage, which allows to fantasize \cite{stanton2021kernel} on new data without re-fitting for more efficient inference.

\subsection{Fail-safe Behaviors}
\label{subsec:fail_safe}
Under-actuated AUVs are generally positively buoyant and therefore rely on their thrusters to maintain depth. This ultimately means that they must maintain a minimum velocity and cannot slow down or stop for replanning. Furthermore, such actions are detrimental to the goal of optimizing area surveyed over time.
Thus, a requirement for real-life performance of our receding horizon IPP framework is to guarantee that it can provide a new path online before reaching the end of the current one to keep the AUV moving.

The presented non-myopic planner can fail to produce valid candidate locations online under two circumstances: i) the transition time $t_t$ between viewpoints is shorter than the minimum runtime of the IPP, $t_r$. This occurs when the distance between $s_i$ and $s_{i+1}$ is too small at the current velocity; ii) the BO procedure, which relies on gradient based optimization with random starts does not converge on time, which can be due to the optimization landscape within the planner horizon being too flat.
To counteract both issues and keep the vehicle in motion, two fail-safes actions are taken, in the following order: Myopic BO if the optimization converged but the rest of the algorithm could not finish on time, or random location selection within a given horizon if the BO optimization could not be carried out either.



\begin{figure}[h!]
    \centering
    \includegraphics[width=\linewidth]{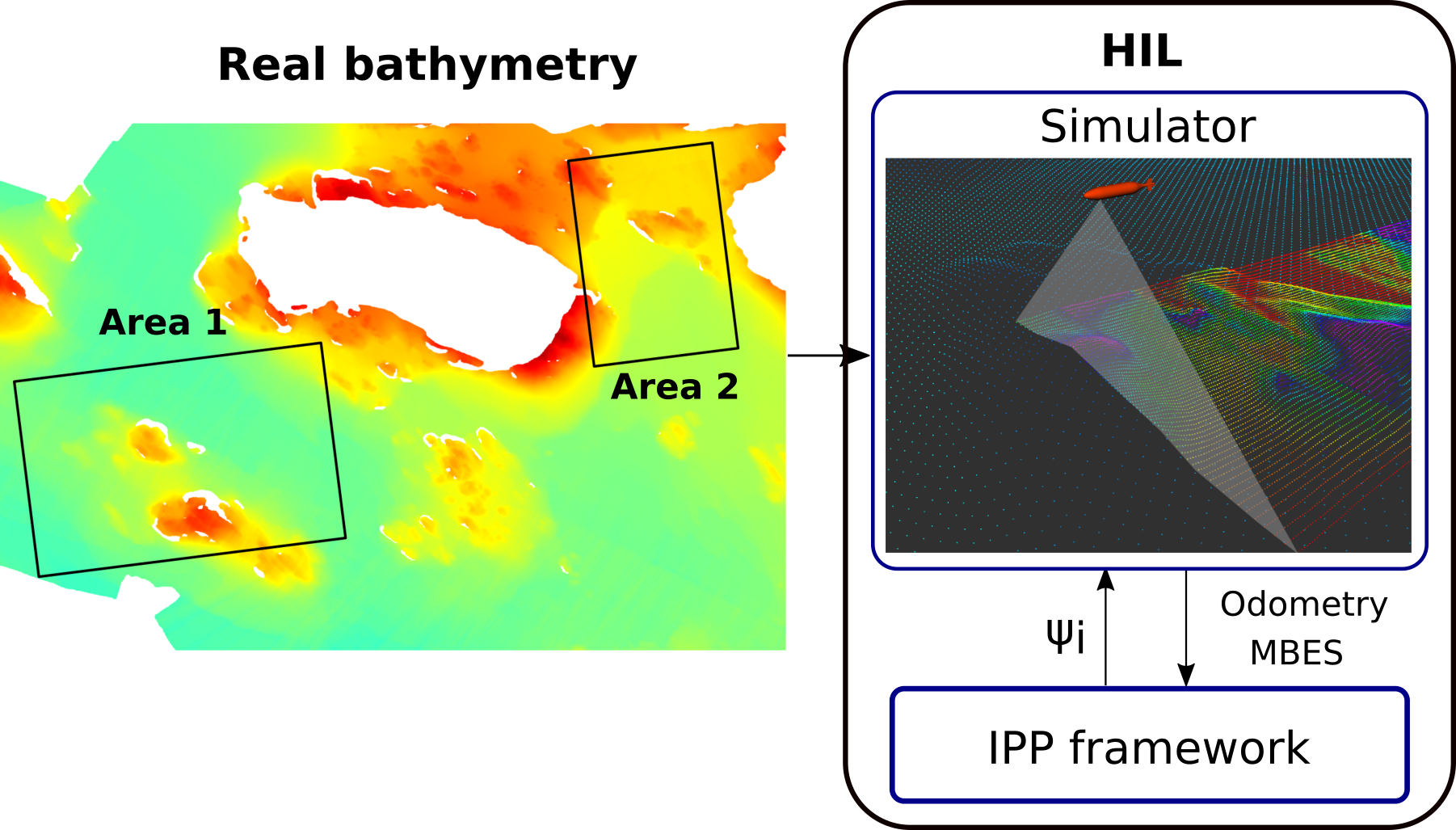}
    \caption{HIL setup in a Jetson Orin: simulated MBES data is generated from real bathymetry with an AUV equipped with a sonar model. Our IPP receives real time odometry and MBES data from the simulator and outputs paths.}
    \label{fig:sim}
\end{figure}

\section{EXPERIMENTAL SETUP}

\subsection{Simulation Environment}
\label{subsec:sim_env}
We assess the performance of the presented framework in the tasks of autonomous bathymetric exploration and mapping of open seabed areas through a set of hardware in the loop experiments in simulation. 
We employ the ROS-based \cite{quigley2009ros} simulation environment from \cite{torroba2022fully} to replicate the bathymetric surveys. 
The underlying bathymetry $E$ from which the simulated MBES are generated was collected off the coast of Sweden by the industrial surveying company Ocean Infinity. The AUV has been parameterized following the hardware constraints of the real AUV Lolo \cite{deutsch2018design}, with a constant surveying velocity of 0.8 $m/s$ and a minimum turning radius of 10m. The AUV is equipped with a simulated MBES with an opening angle of 90$\degree$, a ping rate of 20Hz and 64 beams per ping. To replicate the hardware constraints of the real AUV, the experiments have been carried out in an Nvidia Jetson Orin AGX 64, the payload computer in Lolo. The interface between the IPP framework and the simulated AUV mimics that of the real AUV, used in the experiment with Lolo in \cite{torroba2023online}. The experimental pipeline can be seen in Fig. \ref{fig:sim}.


\subsection{Experiments Configuration}
The two seabed areas used in the experiment can be seen highlighted in Fig. \ref{fig:sim}, and comprise 5 and 10 hectares respectively. These areas were chosen as they contained highly salient features and also large regions of flat terrain. $N_T$ equals 3.6 and 5.8 million MBES beams for surveys $1$ and $2$ respectively, with $T$ being the full duration of the surveys in simulation. The SVGP parameters for both areas can be seen in Table \ref{tab:params}. The kernel used is a Matern.

The Bayesian Optimization stages have been implemented in BoTorch \cite{balandat2020botorch}. The values of the qUCB acquisition function for the experiments can be seen in Table \ref{tab:params}. $q$ has been selected to balance expressiveness of the MC tree and embedded performance. The same reasoning was applied when determining the maximum tree depth $d_{max}$.

\begin{table}[h!]
    \centering
    \renewcommand{\arraystretch}{1.1}
    \setlength{\tabcolsep}{5pt}
     \caption{Method parameters}
    \begin{tabular}{c c c}
        \toprule
        Parameter name & Symbol & Used value\\
        \midrule
        SVGP Kernel value & $\nu$ & 2.5 \\
        SVGP Inducing points & $u$ & 250 \\
        SVGP Mini batch size & $M$ & 1024 \\
        UCB value & $\beta$ & 100 \\  
        UCT value & $c$ & 12  \\
        POMDP discount factor & $\gamma$ & 0.9  \\
        TREE max depth & $d_{max}$ & 2 \\
        TREE branching factor & $q$ & 3 \\
        
        \bottomrule
    \end{tabular}
    \label{tab:params}
\end{table}

\subsection{Performance evaluation}
Ultimately, the goal of the presented IPP framework is to increase mapping efficiency, whose metric can be defined as the quality of the map of an area produced against distance travelled at a constant velocity. We assess the map quality in terms of the difference between the produced map against the ground truth (GT) bathymetry, available in simulation. This consistency error is computed as the difference between the raw GT point cloud and the SVGP posterior sampled at the same resolution than the GT, measured following the procedure presented in \cite{roman2006consistency}. The root mean square of this error (RMSE) against distance travelled is therefore used. 

State of the art methods in IPP with BO \cite{morere2018continuous} and UI \cite{popovic2020informative} use exact GPs in their method formulation. Even for our smaller dataset, retaining the exact GP matrix would require almost two petabytes of memory. Attempting to recreate their methods as baselines in bathymetric surveying is therefore not feasible. 

As suggested in \cite{popovic2024learning,galceran2013survey}, we select the lawn-mowing pattern as our baseline both due to its ubiquity in the surveying industry and the uniformity of the data it provides, which should result in a gold standard SVGP map. We also evaluate against a myopic formulation of our method, using BO with single candidate UCB and without the search tree.

Furthermore, we evaluate of our framework regarding real-time performance on an embedded platform as a mean to gauge its applicability to real world scenarios. To this end, all the HIL experiments have been executed mimicking the parameters of Lolo, as introduced in Sec. \ref{subsec:sim_env}.


\begin{figure}[!t]
    \centering
    \includegraphics[width=1\linewidth]{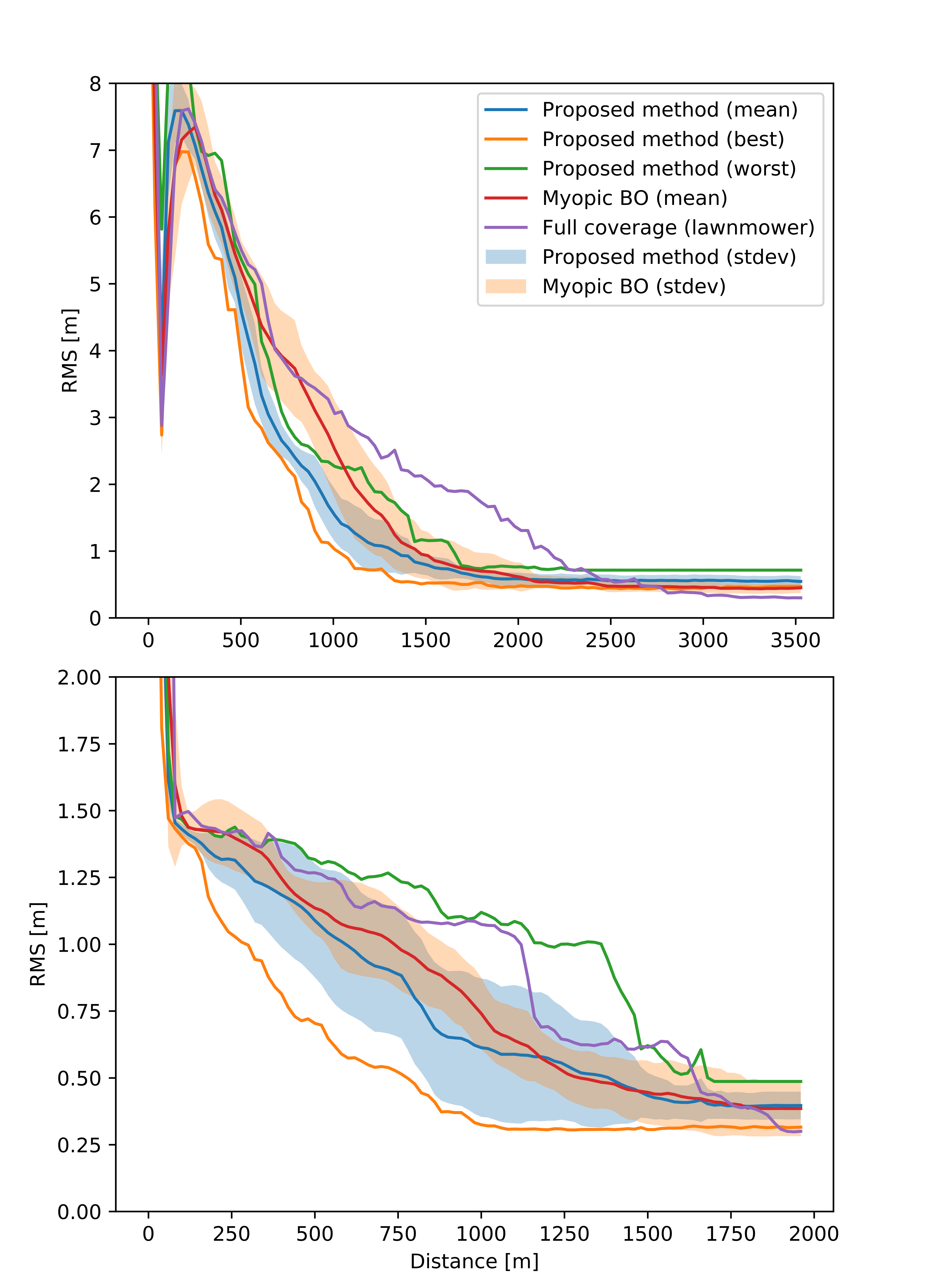}
    \caption{RMS consistency errors over distance of the SVGP-UI maps posteriors against the ground truth bathymetry. The SVGP-UIs regressed with our method converge faster on average (blue) than the lawn-mowing pattern ones (purple) and myopic method (red).}
    \label{fig:results_plots}
\end{figure}

\begin{figure*}[!t]
    \centering
    \includegraphics[width=0.9\linewidth]{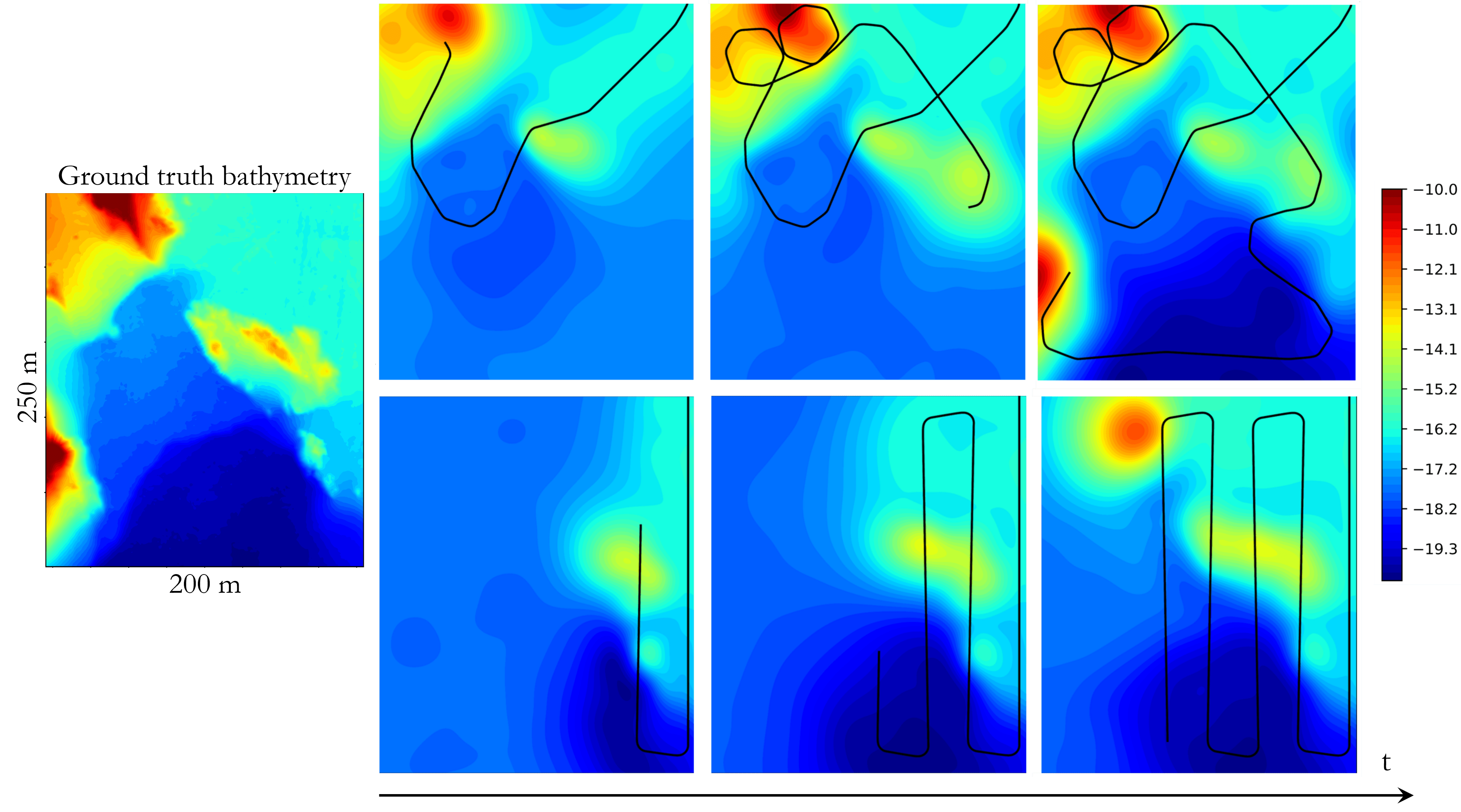}
    \caption{Time lapse of a survey run over area 2 with the AUV path (black) over the SVGP-UI maps posteriors at three time steps. Our IPP has fully mapped the area while the LM has completed $60\%$ of the path. Original bathymetry on the left.}
    \label{fig:bo_vs_lm}
\end{figure*}

\section{Results}

\subsection{Bathymetric surveying}
For each bathymetric area to survey, the results of 10 runs of our IPP method have been averaged and plotted in Fig. \ref{fig:results_plots} against a myopic version relying on single candidate BO, and a lawn-mowing survey with full coverage and minimum overlap between consecutive swaths of 10$\%$. It can be seen that in both experiments our method converges on average faster, particularly on the larger dataset, where the average IPP method could have finished the survey after 2000m, a improvement over distance travelled of 43$\%$. And while the worst-case results remain within the vicinity of the LM performance, it is the best performing cases that best highlight the great potential of the presented algorithm, with a remarkable reduction of survey time of almost 50$\%$.
Additionally, while it must be noted that the IPP residual RMSE is larger than that of the LM pattern in both datasets, the differences are only of 1.2$\%$ and 1.6$\%$ over the initial RMS errors. A time lapse of a run of our IPP (top) against the LM solution can be seen in Fig. \ref{fig:bo_vs_lm}, with the AUV trajectories (black) plotted over the SVGP-UI maps posteriors. It can be easily appreciated how our IPP has mapped the main features while the LM pattern has only covered $~60\%$ of the area.

\subsection{Real time performance}
Regarding real-time performance, Table \cref{tab:timing} presents the average timings of $100$ iterations of our IPP method in the Jetson Orin with the parameters in Table \ref{tab:params}. The average minimum runtime for a successful iteration, $t_r$, has therefore been calculated as the sum of the averages in the table $t_r = 15s$. At the set AUV speed of 0.8 $m/s$, this means that only paths shorter than 12m will prevent the system from running a full iteration and thus result in the fallback behavior in Sec. \ref{subsec:fail_safe}. Considering that 12m is $1.6\%$ of the distance travelled on the smallest survey, we can conclude that the performance of the algorithm is suitable for real-world deployment. 

\begin{table}[h!]
    \centering
     \caption{Average timing results}
    \begin{tabular}{c c}
        \toprule
        Operation & Average time (s) \\
        \midrule
        Optimize $q$ candidates & $3.5\pm 1.2$ \\
        Training $q$ SVGPs & $10.7\pm 4.3$ \\
        Optimize heading & $0.8\pm 0.2$ \\
        \bottomrule
    \end{tabular}
    \label{tab:timing}
\end{table}

From the timing results, it can be stated that the cost of the heading optimization is negligible (and also fixed, as it does not scale with $q$) and we can expect to fully expand a tree of depth  $d=2$ with branching factor $q=3$ most times. However, if the tree is not fully expanded this does not pose a problem as the anytime nature of our tree search will always return the current optimal candidate.

Additionally, while the optimization via the qUCB is slower than single candidate optimization, note that the application of qUCB to discretize the continuous action space into $q$ candidates is still much more performant than training $q$ SVGP maps. As such, there is a large reduction on computational requirements by training fewer maps than what would have been required to condition classic BO on.


\section{Conclusion}
We have presented a non-myopic, double-layered Bayesian optimization framework for IPP aimed at bathymetric exploration and surveying with AUVs equipped with MBES. Our framework produces optimal vehicle paths that maximize information gathered along the paths with a downward looking MBES while abiding to the vehicle motion constraints and regressing accurate online surrogate models of large bathymetric areas with millions of points.

Regarding bathymetric mapping performance, the evaluation of our approach against the standard lawn-mowing pattern and a myopic method highlights its usability particularly in the task of feature search, in which finding informative regions quickly is prioritized over uniform mapping quality. 
The results presented hint at potential reductions of surveying time of $40$-$50\%$, at the cost of residual errors in the final maps $1\%$-$2\%$ higher. Compared to the myopic planner, we expect a larger difference in performance will arise in larger datasets, where short-sightedness penalizes more.

In terms of scalability, the combination of the work in \cite{torroba2022fully, torroba2023online} to produce online SVGP-UI maps, together with the application of batch acquisition functions to non-myopic BO, has enabled our framework to handle surveys orders of magnitude larger than those in \cite{morere2018continuous, popovic2020informative}.

Additionally, while limited to simulation experiments, the runtime performance of our method in the HIL setup presented indicates its amenability to AUV deployment in terms of real-time operability with limited resources.

Finally, our framework is made available in \footnote{\url{https://github.com/ignaciotb/UWExploration/}}.

\section*{ACKNOWLEDGMENT}

This work was supported in part by Stiftelsen för Strategisk Forskning
(SSF) through the Swedish Maritime Robotics Centre (SMaRC) under Grant
IRC15-0046. The authors thank Ocean Infinity for the bathymetric data provided.


\bibliographystyle{IEEEtran} 
\bibliography{IEEEabrv,mybib}
\end{document}